\pdfoutput=1

\documentclass[11pt]{article}

\usepackage{EMNLP2022}

\usepackage{times}
\usepackage{latexsym}

\usepackage[T1]{fontenc}

\usepackage[utf8]{inputenc}

\usepackage{microtype}

\usepackage{inconsolata}

\usepackage{soul}
\usepackage{url}
\usepackage{caption}
\usepackage{subfig}
\usepackage{graphicx}
\usepackage{amsmath}
\usepackage{amsthm}
\usepackage{booktabs}
\usepackage{algorithm}
\usepackage{algorithmic}
\usepackage{xspace}
\usepackage{bm}
\usepackage{multirow}
\usepackage{verbatim}
\usepackage{stfloats}
\usepackage{amssymb}
\usepackage{color}
\usepackage{makecell}
\usepackage{svg}
\newtheorem{theorem}{Theorem}

\usepackage{dcolumn}
\newcolumntype{d}[1]{D{.}{.}{#1}}

\newcommand{\eat}[1]{}
\newcommand{\paratitle}[1]{\vspace{1ex}\noindent \textbf{#1}}

\let\oldhat\hat
\renewcommand{\vec}[1]{\mathbf{#1}}
\renewcommand{\hat}[1]{\oldhat{\mathbf{#1}}}
\renewcommand{\matrix}[1]{\mathbf{#1}}

\newcommand{\wrt}{\emph{w.r.t.}\xspace}
\newcommand{\ie}{\emph{i.e.,}\xspace}

\title{HCL-TAT: A Hybrid Contrastive Learning Method for Few-shot Event Detection with Task-Adaptive Threshold}

\author{Ruihan Zhang$^{1,2}$, Wei Wei$^{1,2}$\thanks{\hspace{0.15cm}Corresponding author: Wei Wei.}, Xian-Ling Mao$^{3}$, Rui Fang$^{4}$, Dangyang Chen$^{4}$\\
        \textsuperscript{1}Cognitive Computing and Intelligent Information Processing (CCIIP) Laboratory,\\
        School of Computer Science and Technology, Huazhong University of Science and Technology\\
        \textsuperscript{2}Joint Laboratory of HUST and Pingan Property \& Casualty Research (HPL)\\
        \textsuperscript{3}Department of Computer Science and Technology, Beijing Institute of Technology\\
        \textsuperscript{4}Ping An Property \& Casualty Insurance company of China, Ltd\\
        \texttt{ruihanzhang@hust.edu.cn}, \texttt{weiw@hust.edu.cn}, \texttt{maoxl@bit.edu.cn}\\
        \texttt{fangrui051@pingan.com.cn, chendangyang273@pingan.com.cn}}

\begin{document}

\maketitle
\begin{abstract}
  
Conventional event detection models under supervised learning settings suffer from the inability of transfer to newly-emerged event types owing to lack of sufficient annotations. A commonly-adapted solution is to follow a identify-then-classify manner, which first identifies the triggers and then converts the classification task via a few-shot learning paradigm. However, these methods still fall far short of expectations due to:  (i) insufficient learning of discriminative representations in low-resource scenarios, and (ii) trigger misidentification caused by the overlap of the learned representations of triggers and non-triggers. To address the problems, in this paper, we propose a novel \textbf{\underline{H}}ybrid \textbf{\underline{C}}ontrastive \textbf{\underline{L}}earning method with a \textbf{\underline{T}}ask-\textbf{\underline{A}}daptive \textbf{\underline{T}}hreshold (abbreviated as HCL-TAT), which enables discriminative representation learning with a two-view contrastive loss (\textit{support-support} and \textit{prototype-query}), and devises an easily-adapted threshold to alleviate misidentification of triggers. Extensive experiments on the benchmark dataset FewEvent demonstrate the superiority of our method to achieve better results compared to the state-of-the-arts. All the code and data of this paper will be available for online public access.

\end{abstract} 

\section{Introduction\label{Sec: 1}}

Event detection (ED) is the subtask of information extraction (IE)~\cite{pan-etal-2021-context-aware,pan-etal-2021-noisy}, which aims at extracting events of task-specified types from an input text and is crucial for many downstream applications such as text summarization \cite{ge2016news} and machine reading comprehension \cite{qiu2019machine}. For example, in the sentence ``Kenyan police intensify manhunt for \textbf{\textit{terror}} suspects", an ideal ED model is to identify ``\textbf{\textit{terror}}'' as a trigger word and classify it into ``\textbf{Conflict.Attack}'' event type. Previous works commonly formulate ED as a token-level classification problem and follow a supervised manner~\cite{chen2015event,liu2017exploiting,nguyen2018graph,zhao2018document,wang-etal-2019-adversarial-training,tong2020improving}. Despite the promising results, in real-world scenarios, new event types are constantly emerging with only a few samples, which gives rise to the problem of low generation on newly-emerged event types, in view of insufficient annotated data. Therefore, it is reasonable to convert the traditional supervised-based event detection task to the few-shot event detection (FSED) problem. 


\begin{figure}[t]
	\centering
	\subfloat[PA-CRF]{\label{fig:visual_a}\includegraphics[width=0.5\linewidth]{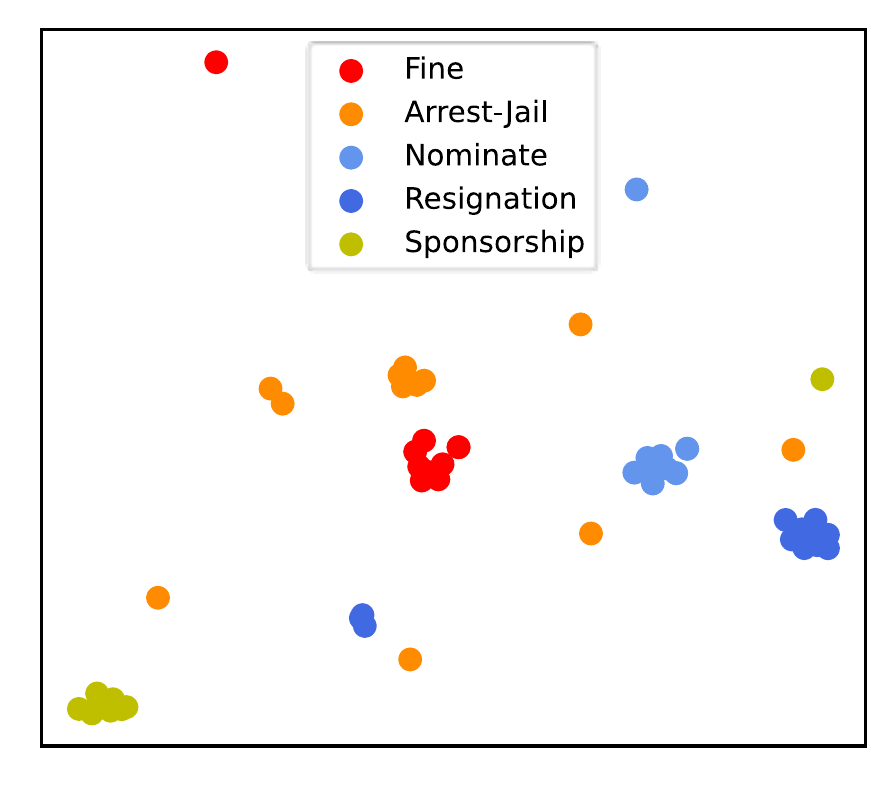}}
	\subfloat[HCL-TAT]{\label{fig:visual_b}\includegraphics[width=0.5\linewidth]{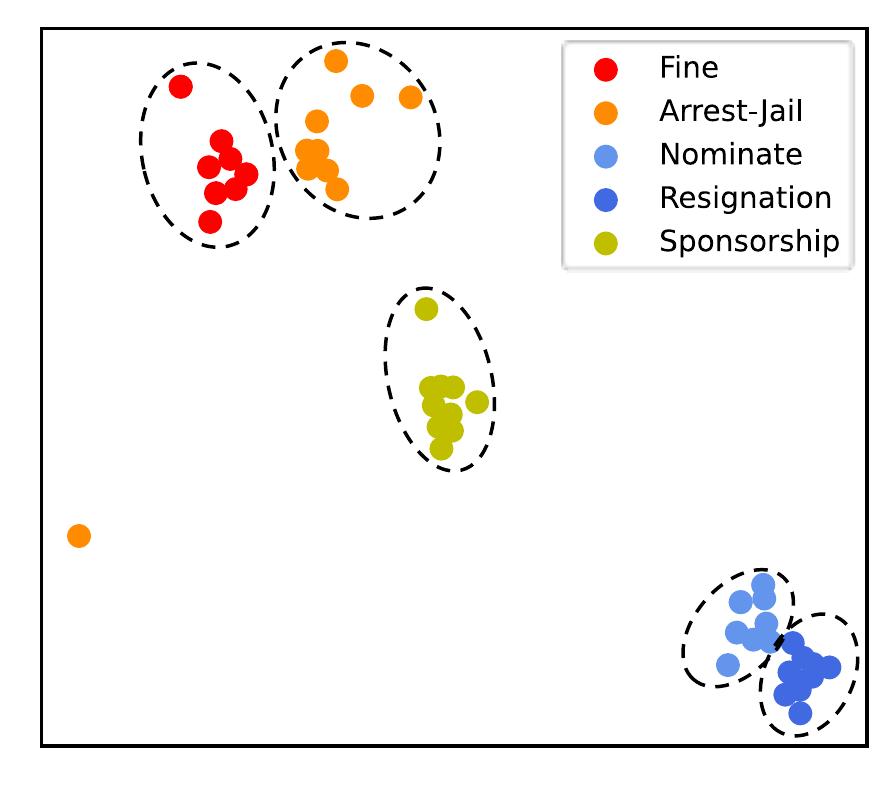}}\\	
	\caption{Visualization of triggers in the same episode on FewEvent test set. The left and right half shows support set representations without and with hybrid contrastive learning, respectively.}
	\label{fig:visual}
\end{figure}

Most works in FSED exploit metric-based meta-learning methods, in which the model needs to learn meta-knowledge from only a few instances in the \textit{support set} and generalize to predict labels for instances in the \textit{query set}. Specifically, in each episode, these methods first build a prototype for each class over the support set, then predict the labels for each instance in query set by matching with the closest prototype representation in the metric space. Indeed, many of these methods follow an identify-to-classify paradigm and are capable of keeping high generation on new event type based on a few observed samples~\cite{deng2020meta,lai2020exploiting,lai2020extensively}. However, the performance of such two-stage model is limited to the problem of error propagation caused by trigger misidentification. Thus, recent works formulate FSED as a few-shot sequence labeling task and apply Conditional Random Field based (CRF-based) methods to jointly identify and classify triggers. Nevertheless, it is usually challenging for these CRF-based methods to fully learn event type dependencies from a single sentence, since a sentence commonly contains only one trigger, or even worse with limited data in FSED. An example is presented in Figure~\ref{fig:visual_a}, even if the state-of-the-art method PA-CRF~\cite{cong2021PACRF} is powerful in exploring event type dependencies via Gaussian distribution for approximation, it is still incapable of generating well discriminative  representations for each event type in the whole embedding space.


Actually, it is non-trivial to solve the problem. Typical metric-based models are mostly optimized with a query-anchored cross-entropy loss based on the similarity between each query instance and all prototypes. However, we find that the gradient directions of prototypes are not always optimized very well during training, and may not guarantee the learnt representations of prototypes are discriminative via optimizing the query-anchored loss, which in turn harms the learning of query instances. Detailed proof can be found in Appendix~\ref{sec:D}.

To address the problem, in this paper, we propose a \textbf{\underline{H}}ybrid \textbf{\underline{C}}ontrastive \textbf{\underline{L}}earning (\textbf{HCL}) method to improve the representations for instances in both support set and query set. Specifically, we first propose a \textbf{\underline{S}}upport-\textbf{\underline{S}}upport \textbf{\underline{C}}ontrastive \textbf{\underline{L}}earning (\textbf{SSCL}) method to makes it prone to generating more discriminative representations of prototypes, via encouraging instances of the same type closer and others of different types farther away within support set. However, the optimization direction of SSCL is still unclear and only information within the support set is considered. Therefore, inspired by~\cite{gao2021contrastive}, we further design a \textbf{\underline{P}}rototype-\textbf{\underline{Q}}uery \textbf{\underline{C}}ontrastive \textbf{\underline{L}}earning (\textbf{PQCL}) method, which pulls query instances together with the prototype of the same type and pushes them apart from prototypes of different types. Logically, PQCL is capable of guiding the learning process of SSCL in a theoretically sound way, and such two contrastive learning losses are helpful for query-anchored loss to generate more discriminative representations for both the support set and the query set. 

Additionally, another serious problem for typical FSED is that most of the words in ED sentences are non-triggers, \ie they do not belong to any event type, called ``O'' type. The unbalanced nature influences the representation learning of triggers, and leads to representation overlap problem between triggers and non-triggers. To address this issue, we further design a \textbf{T}ask-\textbf{A}daptive \textbf{T}hreshold (\textbf{TAT}). Our key insight is that, if the similarity between a query instance and all event types are lower than the average similarity between query instances and ``O'' type, then it is most likely to be a non-trigger. Thus, in each episode, we take the average similarity between all query instances and the prototype of ``O'' type as the threshold to filter misidentified triggers. This threshold can be easily adapted to any episode and ensure the generalization ability. As illustrated in Figure~\ref{fig:visual_b}, by combining the above components, our \textbf{HCL-TAT} could learn more discriminative representations and make better predictions, thus boosting the performance.

The contributions of our work are summarized as follows:
\begin{itemize}
  \item We conduct the study of the limitation of query-anchored cross-entropy loss, namely, indistinguishability in embedding space, which harms the metric-based classification between query instances and prototypes.
  
  \item We propose a hybrid contrastive learning framework for the FSED task, which consists of two components, namely, SSCL and PQCL, to jointly improve the learning of instances on both the support set and the query set, which in return enhance the performance of FSED. Besides, we also propose a task-adaptive task-adaptive threshold (TAT) to eliminate the misidentification of trigger words.
  \item Experiments on FewEvent dataset in different settings demonstrate the advantages of our proposed HCL-TAT over various strong baseline methods.
\end{itemize} 
\section{Related Work}


%

\paratitle{Data-driven Event Detection.}
Early ED approaches rely on traditional machine learning models with handcrafted features to extract events \cite{ji2008refining,liao2010using}. With advances in deep learning, many research efforts have
been dedicated to enhancing ED with different neural network architectures
like \textbf{CNN-based} model~\cite{chen2015event,nguyen2016modeling} and \textbf{RNN-based} model~\cite{nguyen2016joint,feng2016language}. Recently, pre-trained language models (PLMs) are adopted to leverage rich information from large-scale corpora~\cite{yang-etal-2019-exploring,tong2020improving,lai-etal-2020-event}. These methods achieve promising results for supervised ED, but they could not adapt well to FSED with limited data.

\paratitle{Few-shot Event Detection.}
 In recent years, there have been several attempts to conduct meta-learning based approaches~\cite{snell2017prototypical,hu2018relation} for FSED. \citet{deng2020meta} proposed the benchmark FSED dataset FewEvent, and designs a dynamic-memory-based prototypical network (DMBPN) to preserve contextual information of event mentions. \citet{lai2020exploiting} and \citet{lai2020extensively} proposed intra and inter-cluster matching losses to provide more training signals. However, these works categorize events at sentence-level and are solely for the event classification task. Recently, \citet{cong2021PACRF} proposed to treat ED as few-shot sequence labeling task and solved with a modified PA-CRF model, but they still suffer from poor representation learning. In this paper, we propose to exploit contrastive learning to produce more discriminative representations.

\paratitle{Contrastive Learning.}
Contrastive learning has been widely used in various domains based on an idea to pull together similar instances and push apart distinct instances. \textbf{Self-supervised learning} designs specific strategies to generate positive and negative pairs from unsupervised data~\cite{chen2020simple,gao2021simcse,yan-etal-2021-consert,wang-etal-2021-cleve,mcclk2022,KGIC2022,wang2022multi}. As an extension, \textbf{supervised contrastive learning} which leverages label information to generate positive and negative pairs is further proposed~\cite{NEURIPS2020_d89a66c7,gunel2020supervised,wang2021contrastive}. Recently, contrastive learning is also used in few-shot learning tasks to improve the performance~\cite{gao2021contrastive}, by pulling query samples closer to the prototype of the same class and further away from those of different classes. In this paper, we propose a hybrid contrastive learning (HCL) method, composed of two complementary contrastive losses, to generate more discriminative representations for both support and query set. With the help of a task-adaptive threshold (TAT) method, the proposed HCL-TAT can fully make use of limited data and boosts the performance.

\section{Methodology}

\begin{figure*}[t]
    \centering
    \includegraphics[width=1.0\textwidth]{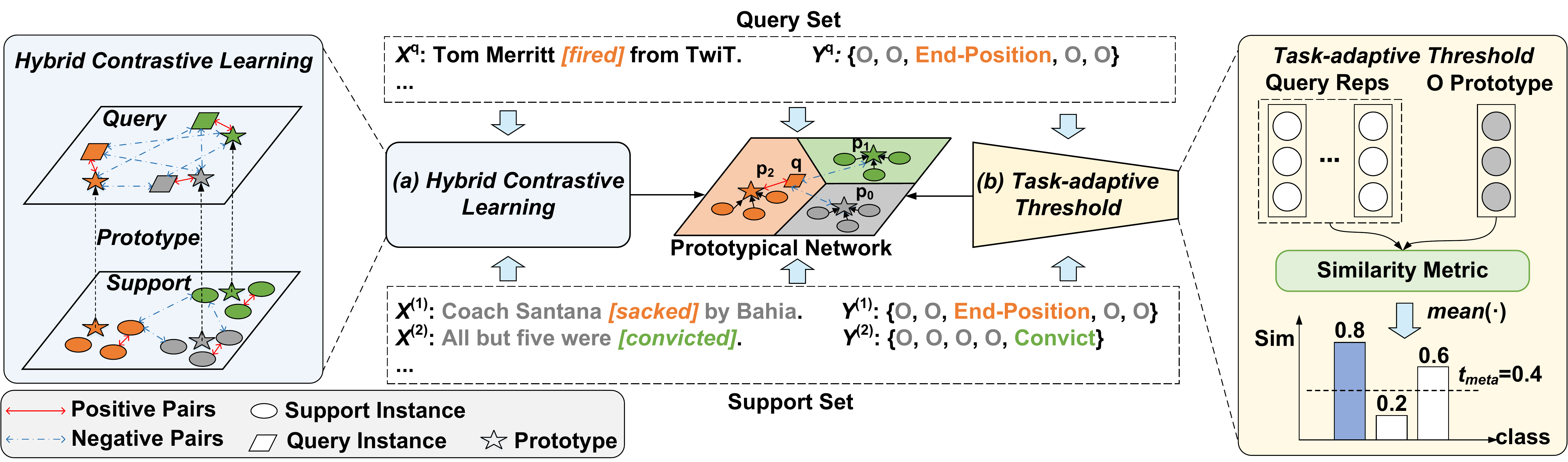}
    \caption{Overall framework of the proposed HCL-TAT model. HCL-TAT is based on a prototypical network, composed of two components: (a) hybrid contrastive learning including support-support contrastive learning and prototype-query contrastive learning; (b) task-adaptive threshold based on the logits in each episode.} 
    \label{fig:model}
\end{figure*}


\subsection{Overview}
An overview of the proposed HCL-TAT method is illustrated in Figure~\ref{fig:model}. We first give the problem statement in Section~\ref{Sec: 3.2}, then introduce the backbone prototypical network in Section~\ref{Sec: 3.3}. In Section~\ref{Sec: 3.4} and Section~\ref{Sec: 3.5}, we detailedly describe the hybrid contrastive learning and task-adaptive threshold method respectively. Finally, the training process is given in Section~\ref{Sec: 3.6}.

%
%

\subsection{Problem Statement\label{Sec: 3.2}}

In this paper, we formulate few-shot event detection (FSED) as a few-shot sequence labeling task. Given a sentence $\mathcal{X}=\{x_1, x_2, ..., x_n\}$ composed of $n$ tokens and its corresponding label sequence $\mathcal{Y}=\{y_1, y_2, ..., y_n\}$, each token $x_i$ is categorized into an event type. Besides the defined event types in the dataset, we add an ``O'' type to denote for tokens that do not belong to any event type (called non-triggers). Then, FSED is defined with a typical $N$-way-$K$-shot setting. Specifically, given a support set $\mathcal{S}=\{\mathcal{X}^{(i)}, \mathcal{Y}^{(i)}\}_{i=1}^{N \times K}$ which has $N$ event types with $K$ labeled samples for each type, and a query set $\mathcal{Q}=\{\mathcal{X}^{(i)}, \mathcal{Y}^{(i)}\}_{i=1}^{N \times M}$ which has the same $N$ event types as $\mathcal{S}$ with $M$ samples for each type, we formulate a $N$-way-$K$-shot task or episode as $\mathcal{T}=\{\mathcal{S}, \mathcal{Q}\}$. Note that in each episode we also have an additional ``O'' type. The goal of FSED is to predict the labels for instances in $\mathcal{Q}$ based on $\mathcal{S}$. Specifically, in training stage, the classification results of $\mathcal{Q}$ are used to update model parameters, while in testing stage, the classification results are used to evaluate the model. The training phase is composed of a set of episodes $\mathcal{T}_{train}=\{\mathcal{T}_i\}_{i=1}^{M_{train}}$, and the testing phase is composed of another set of episodes $\mathcal{T}_{test}=\{\mathcal{T}_i\}_{i=1}^{M_{test}}$. $M_{train}$ and $M_{test}$ denote the number of training episodes and testing episodes respectively, and the label set of $\mathcal{T}_{train}$ and $\mathcal{T}_{test}$ are disjoint.

\subsection{Prototypical Network\label{Sec: 3.3}}


Following~\citet{cong2021PACRF}, we select BERT as the text encoder to represent event mentions. We encode the sentence $\mathcal{X}$ into a sequence of hidden embeddings as follows,
\begin{equation}
    \label{equ:sentence_encoder}
    \{\vec{h}_1, \vec{h}_2, ..., \vec{h}_n\} = f(\mathcal{X},\theta),
\end{equation}
where $f(:,\theta)$ is the encoder and $\vec{h}_i$ is the hidden representation of each token $x_i$.

We then adopt the widely-used prototypical network (Proto)~\cite{snell2017prototypical} as our backbone, in which the classification results are obtained by matching instances with prototypes of each class in the metric space. During each episode, Proto first computes prototypes for each class $c$ by averaging instances in support set:
\begin{equation}
    \label{eq:proto_compute}
    \vec{p}_c = \frac{1}{K}\sum_{i \in \mathcal{S}(c)}\vec{h}_i, \;c=0,1,...,N,
\end{equation}
where $\vec{p}_i$ is the prototype for class $c$, $\mathcal{S}_{c}$ represents for all the words of class $c$ in $\mathcal{S}$. Then in training phase, the prototypes and query instances are fed into a non-parametric distance-based classifier to compute the cross-entropy loss for few-shot classification:
\begin{gather}
    \label{eq:ce_loss}
    \mathcal{L}_{CE} = -\sum_{(x_i,y_i) \in \mathcal{Q}}\operatorname{log}P(y_i|x_i,\mathcal{S}),\\
    P(y_i|x_i,\mathcal{S}) = \frac{\operatorname{exp}(-d(\vec{h}_i, \vec{p}_{y_i}))}{\sum_{c \in \mathcal{C}}\operatorname{exp}(-d(\vec{h}_i, \vec{p}_{c}))},
\end{gather}
where $\mathcal{C} = \{0,1,...,N\}$ denotes for the sampled class set including an additional ``O'' type, $d(\cdot)$ is a metric function to measure the similarity between query instances and prototypes.

We then analyze the bottleneck of query-anchored loss in Eq. (\ref{eq:ce_loss}). Assuming $d(\cdot)$ is dot product, $\vec{p}^{pos}$ represents for the prototype of class $y_i$ and $\vec{p}^{n}$ represents for any prototype of different classes, we compute the gradient \wrt $\vec{h}_i$, $\vec{p}^{n}$ and $\vec{p}^{pos}$ respectively:
\begin{gather}
    \frac{\partial \mathcal{L}_{CE}}{\partial \vec{h}_i} = \frac{\sum_{n}\Delta_n (\vec{p}^n - \vec{p}^{pos})}{1 + \sum_{n}\Delta_n},\\
    \frac{\partial \mathcal{L}_{CE}}{\partial \vec{p}^{n}} = \frac{\Delta_n \vec{h}_{i}}{1 + \sum_{n}\Delta_n},
    \frac{\partial \mathcal{L}_{CE}}{\partial \vec{p}^{pos}} = -\frac{\sum_{n}\Delta_n \vec{h}_{i}}{1 + \sum_{n}\Delta_n},\\
    \Delta_n = \operatorname{exp(\vec{h}_i \cdot \vec{p}^n - \vec{h}_i \cdot \vec{p}^{pos})}.
\end{gather}

We can conclude that the query instance $\vec{h}_i$ has a better update direction compared with prototypes, and this makes it hard to learn good representations for the prototypes, which could in turn affects the learning of $\vec{h}_i$, since the gradient \wrt $\vec{h}_i$ approaches to zero when $\vec{p}^{pos}$ and $\vec{p}^{n}$ are close in the embedding space. Therefore, we propose hybrid contrastive learning method to learn more discriminative representations and break this bottleneck. Proof details can be found in Appendix~\ref{sec:D}.

\subsection{Hybrid Contrastive Learning\label{Sec: 3.4}}

Hybrid contrastive learning (HCL) consists of two components: support-support contrastive learning (SSCL) applied to instances within $\mathcal{S}$ for more discriminative prototype representations and prototype-query contrastive learning (PQCL) applied between $\mathcal{S}$ and $\mathcal{Q}$ to guide the optimization process for instances in both set.

\subsubsection{Support-Support Contrastive Learning}

To produce better prototypes, we should learn more discriminative representations for instances in $\mathcal{S}$. Therefore, naturally we follow a supervised contrastive learning manner to construct our SSCL loss. We use label information to construct positive and negative pairs. In each episode, for word $x_i$, we take instances of its same class in $\mathcal{S}$ as positive pairs $\mathcal{P}(i)$ and instances of different classes in $\mathcal{S}$ as negative pairs $\mathcal{N}(i)$. Empirically, since a non-linear projection layer improves the representation quality for contrastive learning~\cite{chen2020simple}, we exploit a 2-layer MLP to project the token representations to a latent space.
\begin{equation}
    \label{projection_head}
    \tilde{\vec{h}}_i = \matrix{W}_2\sigma(\matrix{W}_1\vec{h}_i),
\end{equation}
where $\matrix{W}_1$ and $\matrix{W}_2$ are trainable weights, and $\sigma$ is an activation function.

We then encourage the positive pairs to pull closer and negative pairs to push further. In this way, we make instances in $\mathcal{S}$ more discriminative and produce more compact clusters to generate better prototypes. Specifically, we take dot product to measure the similarity between instances, and the SSCL loss is calculated as follows:
\begin{gather}
    \label{equ:SSCL_loss}
    \mathcal{L}_{SSCL} = \sum_{(x_i,y_i) \in \mathcal{S}}\mathcal{L}_{SSCL_{i}},\\
    \mathcal{L}_{SSCL_{i}} = -\operatorname{log}\frac{\operatorname{exp}(\tilde{\vec{h}}_i \cdot \tilde{\vec{h}}_j / \tau)}{\sum_{k \neq i}\operatorname{exp}(\tilde{\vec{h}}_i \cdot \tilde{\vec{h}}_k / \tau)},
\end{gather}
where $\tau$ is a scalar temperature parameter. In practice, we apply $\ell_2$ normalization to $\tilde{\vec{h}}_i$ for numerical stability.

\subsubsection{Prototype-Query Contrastive Learning}

By adopting SSCL, instances in the support set are enforced to be more separable. However, the optimization direction is still unclear without the guidance of query set. To this end, inspired by~\cite{gao2021contrastive}, we propose prototype-query contrastive learning (PQCL) to eliminate the gap. Similarly, we take each prototype as the anchor, and regard query instances of the same class as positive pairs and query instances of different classes as negative pairs. Given a class $c \in \mathcal{C}$, the positive pairs set and negative pairs set are defined as $\mathcal{Q}_{c}^{pos}$ and $\mathcal{Q}_{c}^{neg}$ respectively. The PQCL loss is calculated by:
\begin{gather}
    \label{equ:PQCL_loss}
    \mathcal{L}_{PQCL} = \sum_{c \in \mathcal{C}}\sum_{(x_i, y_i) \in \mathcal{Q}_{c}^{pos}}\mathcal{L}_{PQCL_{c}^{i}},\\
    \mathcal{L}_{PQCL_{c}^{i}} = -\operatorname{log}\frac{sim_{c}^{i}}{sim_{c}^{i} + \sum_{(x_k, y_k) \in \mathcal{Q}_{c}^{neg}}sim_{c}^{k}},\\
    sim_{c}^{i} = \operatorname{exp}(\vec{p}_{c} \cdot \tilde{\vec{h}}_i / \tau).
\end{gather}

We then combine the two contrastive losses as a hybrid contrastive learning (HCL) method, to produce more discriminative prototypes for query instances. Our HCL is complementary to the query-anchored cross-entropy loss $\mathcal{L}_{CE}$, and by optimizing the losses jointly, the model can obtain better representations for both support set and query set.

\subsection{Task-adaptive Threshold\label{Sec: 3.5}}

The proposed HCL could produce better representations for both support set and query set. However, since ``O'' type contains all instances that do not belong to any event type, the semantics of those instances are complex, making it hard to separate them even with the help of HCL. Therefore, it's important to further regularize the classification process for more accurate predictions. Naturally, we can set a threshold and let query instances whose similarity to all event types are lower than the threshold to output an ``O'' type prediction.

However, manually designed thresholds are hard to tune and lack generalization. Thus, based on the episodic training in FSED, we propose a task-adaptive threshold method. Intuitively, since the similarity between query instance and prototype can be regarded as a confidence score, in each episode, we take the average value of similarities between all query instances and ``O'' type as the threshold for automatically separating triggers and non-triggers. If the similarity between a query instance and an event type cannot even reach the calculated threshold, then it's unlikely for the query instance to be a trigger of that event type. We use probability to represent for similarity and the threshold $t$ can be defined as:
\begin{equation}
    \label{eq:threshold}
    t_{meta} = \frac{1}{|\mathcal{Q}|}\sum_{(x_i, y_i) \in \mathcal{Q}}P(y_i = 0 | x_i, \mathcal{S}).
\end{equation}

In this way, we could regularize the classification process by eliminating the misidentification of triggers.

\subsection{Training Process\label{Sec: 3.6}}

By combining the hybrid contrastive loss and task-adaptive threshold, we obtain the full method HCL-TAT. In each training episode, the model is optimized with the mixture of query-anchored cross-entropy loss and hybrid contrastive loss:
\begin{equation}
    \label{final_loss}
    \mathcal{L} = \mathcal{L}_{CE} + \alpha \mathcal{L}_{SSCL} + \beta \mathcal{L}_{PQCL},
\end{equation}
where $\alpha$ and $\beta$ are trade-off parameters to balance the losses. During training and testing phase, we both use the threshold to regularize the classification. Finally, we randomly select multiple testing episodes and report the mean result to evaluate the performance.


\section{Experiments}

\subsection{Experimental Settings}

\paragraph{Dataset.} 
We conduct experiments on the newly-proposed largest few-shot event dataset \textbf{FewEvent}\footnote{https://github.com/231sm/Low\_Resource\_KBP} proposed by \cite{deng2020meta}. FewEvent contains more than 70000 event instances over 100 fine-grained event types, and is built from ACE2005, KBP2017 and external knowledge bases like Freebase and Wikipedia, thus it's representative and can take coverage of most event types. Following~\cite{cong2021PACRF}, we use the same 80 event types for training, 10 event types for validation and the rest 10 event types for testing, where event types between each subset are disjoint. The statistics of FewEvent dataset is listed in Appendix~\ref{sec:A}.

\paragraph{Evaluation.}
Same as previous works \cite{chen2015event}, we report micro precision, recall and F1 score over event types, among which F1 score is the most important metric. Following~\cite{cong2021PACRF}, we use the same episodic evaluation method to evaluate our model in few-shot settings, by randomly selecting episodes containing $N$-way-$K$-shot samples from the test set. We run each experiment 5 times to get the averages and standard deviations for fair comparison.

\paragraph{Implementation Details.}
We use bert-base-uncased model through huggingface's transformers\footnote{https://github.com/huggingface/transformers} to obtain the 768-dimensional token-level embeddings. Following~\cite{cong2021PACRF}, we also use episodic training and evaluation methods for few-shot event detection. In the training stage, we randomly select $N$-way-$K$-shot samples from training set for 20000 iterations. In each iteration, we first randomly select $N$ event types then assign $K$ instances for each type as the support set, and select other $M$ samples for each type as the query set. We manually tune the hyper-parameters by running similar $N$-way-$K$-shot episodic evaluation on the validation set for 1000 iterations, and the final test result is obtained by 3000 iterations on the test set. Since only 0.12\% of sentences are longer than 128, we set the maximum length to 128 and pad sentences shorter than 128 with a special character ``[PAD]''. We optimize our model using AdamW with a 1e-5 learning rate. For the contrastive learning part, the scalar temperature parameter $\tau$ is set to 0.5 and 0.1 for SSCL and PQCL, respectively, and the trade-off parameters of $\alpha$ and $\beta$ are both set to 0.5. We experiment with Pytorch 1.10 in Ubuntu 18.04. For 5-way-5-shot, 5-way-10-shot and 10-way-5-shot settings, we run the experiments on one NVIDIA GeForce RTX 2080 Ti. For 10-way-10-shot setting, we run the experiments on one NVIDIA GeForce RTX 3090.

\subsection{Baselines}

\begin{table*}[htb]
	\centering
	\scalebox{1.0}{
	\begin{tabular}{l|cccc}
		\toprule[1pt]
		\textbf{Model}    & \textbf{5-way-5-shot}    & \textbf{5-way-10-shot}    & \textbf{10-way-5-shot} & \textbf{10-way-10-shot}  \\
		\midrule
		LoLoss & 31.51 $\pm$ 1.56 & 31.70 $\pm$ 1.21 & 30.46 $\pm$ 1.38 & 30.32 $\pm$ 0.89 \\
		MatchLoss & 30.44 $\pm$ 0.99 & 30.68 $\pm$ 0.78 & 28.97 $\pm$ 0.61 & 30.05 $\pm$ 0.93 \\
		DMBPN   & 37.51 $\pm$ 2.60 & 38.14 $\pm$ 2.32 & 34.21 $\pm$ 1.45 & 35.31 $\pm$ 1.69 \\
		\midrule
		Proto-dot\dag & 41.54 $\pm$ 3.82 & 42.21 $\pm$ 0.68 & 33.27 $\pm$ 2.37 & 39.23 $\pm$ 2.95 \\
		Match\dag & 30.09 $\pm$ 1.71 & 48.10 $\pm$ 1.38 & 28.94 $\pm$ 1.15 & 45.91 $\pm$ 1.98 \\
		Proto\dag & 47.30 $\pm$ 2.55 & 54.81 $\pm$ 2.27 & 42.48 $\pm$ 1.00 & 50.14 $\pm$ 0.65 \\
		Vanilla CRF & 59.01 $\pm$ 0.81 & 62.21 $\pm$ 1.94 & 56.00 $\pm$ 1.51 & 59.35 $\pm$ 1.09 \\
		CDT & 59.30 $\pm$ 0.23 & 62.77 $\pm$ 0.12 & 56.41 $\pm$ 1.09 & 59.44 $\pm$ 1.83 \\
		PA-CRF & 62.25 $\pm$ 1.42 & 64.45 $\pm$ 0.49 & 58.48 $\pm$ 0.68 & 61.64 $\pm$ 0.81 \\
		\midrule
		\textbf{HCL-TAT}    & \textbf{66.96} $\pm$ 0.70 & \textbf{68.80} $\pm$ 0.85 & \textbf{64.19} $\pm$ 0.96 & \textbf{66.00} $\pm$ 0.81 \\
		\bottomrule[1pt]
	\end{tabular}}
    \caption{F1 scores ($10^{-2}$) of evaluated methods on FewEvent test set. \dag\, means the model is re-implemented by ourselves. The best scores are highlighted in boldface, with $p < 0.02$ under t-test.}
    \label{table_main_result}
\end{table*}

To comprehensively investigate the effectiveness of our method, we choose a variety of baselines on FewEvent, including two-stage methods and unified methods.


For \textbf{two-stage models}, we compare with \textbf{LoLoss}~\cite{lai2020exploiting}, \textbf{MatchLoss}~\cite{lai2020extensively} and \textbf{DMBPN}~\cite{deng2020meta}, by adding a trigger identification model before their methods. LoLoss uses matching information between examples in the support set as additional training signals, MatchLoss extends LoLoss to consider intra-cluster matching and inter-cluster information. DMBPN proposes the benchmark FSED dataset FewEvent, and proposes a dynamic memory network based model to preserve more contextual information for FSED.

For \textbf{unified models}, we compare with three typical few-shot classification baselines, \textbf{Proto-dot}, \textbf{Match} and \textbf{Proto}. Besides, we also compare with the state-of-the-art CRF-based methods, \textbf{Vanilla CRF}, \textbf{CDT}~\cite{hou-etal-2020-shot} and \textbf{PA-CRF}~\cite{cong2021PACRF}. Match, Proto and Proto-dot are similar metric-based model, with cosine similarity, euclidean distance and dot product used as metric respectively. We re-implement the three baselines since Cong ``[PAD]'' embeddings are used to calculate prototype of ``O'' type in previous work~\cite{cong2021PACRF}. Vanilla CRF adds a simple CRF layer behind baseline models, while CDT and PA-CRF exploits Gaussian distribution approximation and collapsed dependency transfer to enhance the CRF layer, respectively.

\subsection{Overall Performance}

We use Proto as the backbone model for HCL-TAT. Table \ref{table_main_result} show the overall performance on FewEvent test set. We can observe that: (i) Compared with two-stage models, unified models achieve much better results, which demonstrates the influence of error propagation and the advantage of unifed architectures. (ii) Among the three metric-based few-shot classification baselines, Proto achieves the best result. This indicates that euclidean distance is the best metric for FSED. Note that this 
observation conflicts with the conclusion in PA-CRF, because the authors uses ``[PAD]'' embeddings to calculate the prototype of ``O'' type. (iii) The re-implemented baselines have a lower performance compared with those in~\cite{cong2021PACRF}. This implies that when optimizing with only query-anchored loss, the model cannot learn discriminative representations and even using ``[PAD]'' embeddings to calculate the prototype of ``O'' type is a better choice than averaging the corresponding instance representations. (iv) Our proposed HCL-TAT achieves new state-of-the-art results under all four settings. Specifically, compared with PA-CRF, HCL-TAT brings an improvement of 4.71\%, 4.35\%, 5.71\% and 4.36\%, which proves the effectiveness of HCL-TAT to learn better representations and obtain better classification results. (v) Our proposed HCL-TAT achieve even higher results in 5-shot settings, which demonstrates the ability of HCL-TAT to make full use of limited data compared with other methods.


\subsection{Ablation Study}

\begin{table*}[htb]
	\centering
	\scalebox{0.94}{
    \begin{tabular}{l|ccc|ccc}
        \toprule
        \multicolumn{1}{l|}{\multirow{2}*{\textbf{Model}}}      & \multicolumn{3}{c|}{\textbf{5-way-5-shot}} & \multicolumn{3}{c}{\textbf{5-way-10-shot}} \\
        \cmidrule{2-7}
        \multicolumn{1}{c|}{} & \textbf{P}       & \textbf{R}       & \textbf{F1}     & \textbf{P}       & \textbf{R}      & \textbf{F1}     \\
        \midrule
        HCL-TAT                 & \textbf{62.63} $\pm$ 2.31    & 72.04 $\pm$ 1.93    & \textbf{66.96} $\pm$ 0.70    & \textbf{63.87} $\pm$ 2.35       & 74.65 $\pm$ 1.36      & \textbf{68.80} $\pm$ 0.85      \\
        \;w/o SSCL                      & 59.61 $\pm$ 2.48       & 71.65 $\pm$ 1.72       & 65.03 $\pm$ 0.82       & 60.22 $\pm$ 4.78       & 74.38 $\pm$ 1.81      & 66.42 $\pm$ 2.34      \\
        \;w/o PQCL                      & 57.50 $\pm$ 1.80    & 71.88 $\pm$ 1.52    & 63.85 $\pm$ 0.67    & 60.88 $\pm$ 2.18    & 72.63 $\pm$ 1.23   & 66.21 $\pm$ 1.14   \\
        \;w/o HCL                      & 49.52 $\pm$ 4.34 & 74.67 $\pm$ 3.36 & 59.38 $\pm$ 2.59  & 57.72 $\pm$ 2.72   & 73.35 $\pm$ 1.10  & 64.57 $\pm$ 1.69  \\
        \;w/o TAT          & 46.69 $\pm$ 1.25       & \textbf{76.98} $\pm$ 0.29       & 58.12 $\pm$ 0.94       & 49.56 $\pm$ 1.11       & 76.33 $\pm$ 0.67      & 60.09 $\pm$ 0.92      \\
        \bottomrule
    \end{tabular}}
    \caption{Precision, recall and F1 scores ($10^{-2}$) of ablation study results on FewEvent test set. When remove both HCL and TAT, the method degenerates to a Proto model.}
    \label{table_ablation_study}
\end{table*}

We conduct ablation studies to investigate the effectiveness of each component in the proposed HCL-TAT model. The experimental results are shown in Table~\ref{table_ablation_study}.

\paragraph{Effect of Contrastive Learning.} We remove the two contrastive losses in HCL to observe the performance. We can conclude that: (i) When removing SSCL, the F1 score drops 1.93\%/2.38\% under 5-way-5-shot and 5-way-10-shot settings respectively, which indicates that SSCL could benefit the model by producing more separable representations for prototypes. (ii) When removing PQCL, the F1 scores drop by 3.11\%/2.59\%. This shows that compared with SSCL, PQCL contributes more to the improvement due to the information interaction between support and query set. Especially, in 5-way-10-shot-setting, the performance gap between SSCL and PQCL is slight, we believe the reason is that more instances in support set provide more training signals for SSCL. (iii) When removing both SSCL and PQCL, \ie HCL, the F1 scores drop significantly by 7.58\%/4.23\%, which proves the effectiveness of combining the two contrastive losses. Besides, we note that HCL achieves higher improvement in more difficult setting (5-shot), which further proves the superiority of HCL in few-shot scenarios. (iv) The improvement of HCL mainly comes from the improvement of precision score, which indicates that by producing more discriminative representations, HCL reduces the representation overlap between different types and thus alleviates the misidentification of triggers.


\paragraph{Effect of Task-adaptive Threshold.} To prove the effect of task-adaptive threshold (TAT), we remove this module and evaluate the performance under 5-way-5-shot and 5-way-10-shot settings. The results show that without TAT, the F1 scores drop dramatically by 8.84\%/8.71\%, in which the precision scores contribute most, with a 15.94\%/14.31\% decrease. This demonstrates that TAT could on the other hand eliminate the misidentification of triggers by regularizing the classification process.

\subsection{Performance of Trigger Identification}

\begin{table}[htb]
	\centering
	\scalebox{1.0}{
	\begin{tabular}{l|cc}
		\toprule[1pt]
		\textbf{Model}    & \textbf{FSTI}    & \textbf{FSED}  \\
		\midrule
		PA-CRF & 63.68 & 62.25  \\
		HCL-TAT & 68.18 & 66.96 \\
		\bottomrule[1pt]
	\end{tabular}}
    \caption{Average F1 scores ($10^{-2}$) of HCL-TAT and PA-CRF on FSTI and FSED tasks, on FewEvent test set under 5-way-5-shot setting.}
    \label{table_trigger_identification_result}
\end{table}

To investigate the effectiveness of our method to solve trigger identification problem, we compare the results between HCL-TAT and the state-of-the-art method PA-CRF. As shown in Table~\ref{table_trigger_identification_result}, HCL-TAT achieves 68.18\% on few-shot trigger identification (FSTI) task, bringing a +4.5\% improvement compared with PA-CRF. This indicates that the main improvement of HCL-TAT comes from more accurate identification of trigger words, which can be attributed to two aspects. First, HCL-TAT learns more discriminative representations for both support set and query set through hybrid contrastive learning. Second, the introduced task-adaptive threshold further regularizes the classification process to avoid misidentification.

\subsection{N-way-K-shot Evaluation}

\begin{figure}[htb]
    \centering
    \includegraphics[width=1.0\linewidth]{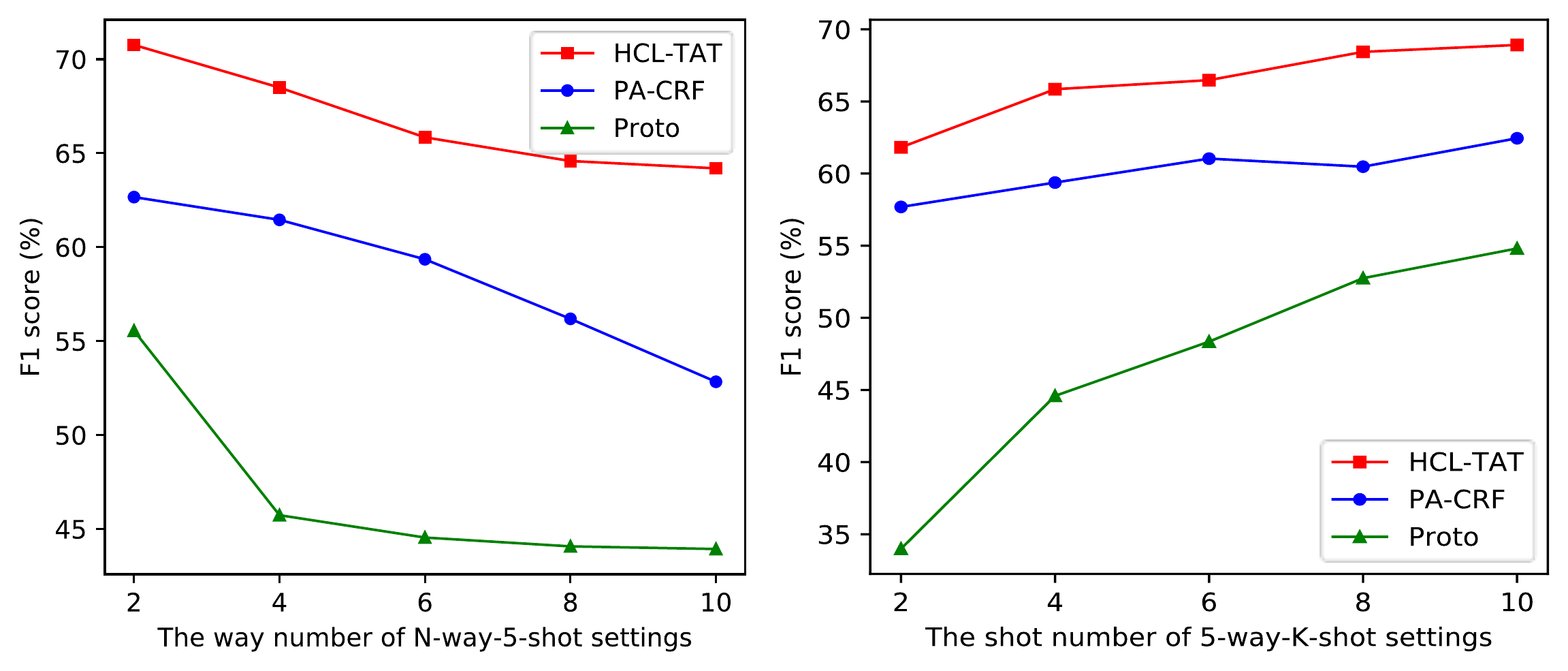}
    \caption{$N$-way-$K$-shot evaluations for three different models. The left part illustrates F1 scores in $N$-way-5-shot settings, and the right part illustrates F1 scores in 5-way-$K$-shot settings. We run each experiment once to analyze the tendency of F1 scores.}
    \label{fig:n-way-k-shot-evaluation}
\end{figure}

We conduct $N$-way-$K$-shot evaluation to investigate the performance tendency of different models in different few-shot settings. For fair comparison, we re-run PA-CRF with the released source code to obtain the results. As illustrated in Figure~\ref{fig:n-way-k-shot-evaluation}, generally, when the shot number $K$ is fixed, F1 scores tend to drop with the increase of way number $N$, and when the way number $N$ is fixed, F1 scores tend to improve with the increase of shot number $K$. Besides, we can observe that the backbone Proto model is more sensitive to the change of way and shot numbers, which indicates that the vanilla prototypical network is more vulnerable to limited data. Moreover, in all settings, HCL-TAT outperforms other methods with a large margin, especially when $N$ is large. This proves that our method shows more robustness to the way number with the help of HCL to fully exploit information between event types and produce more discriminative representations.

\subsection{Learning Visualization}

\begin{figure*}[htb]
	\centering
	\subfloat[Proto]{\label{fig:visual_proto}\includegraphics[width=0.25\linewidth]{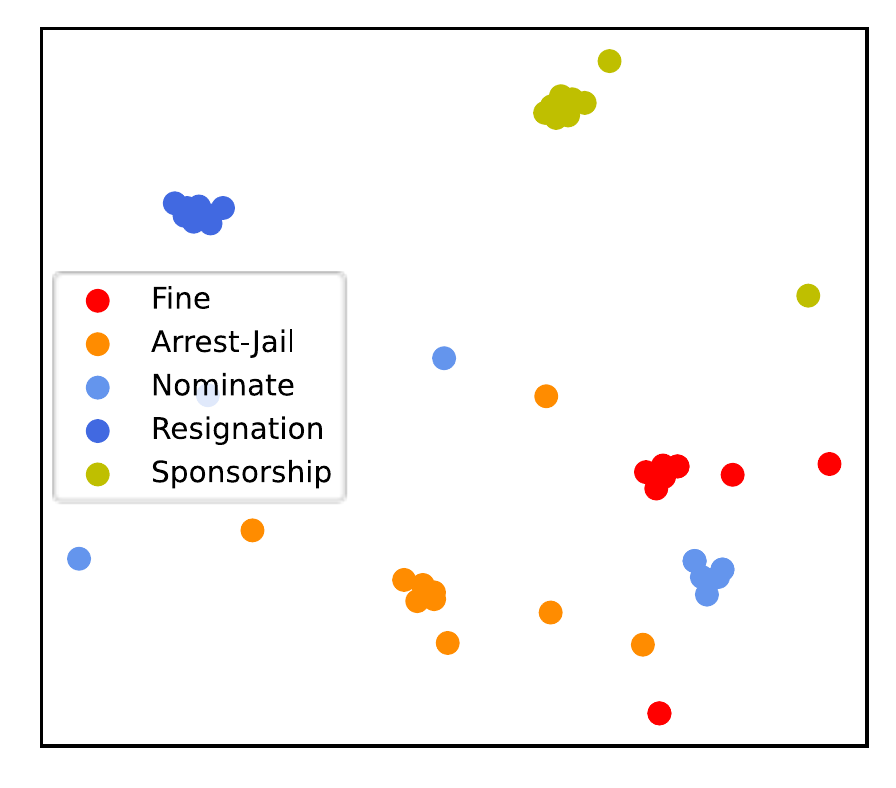}}
	\subfloat[Proto+SSCL]{\label{fig:visual_sscl}\includegraphics[width=0.25\linewidth]{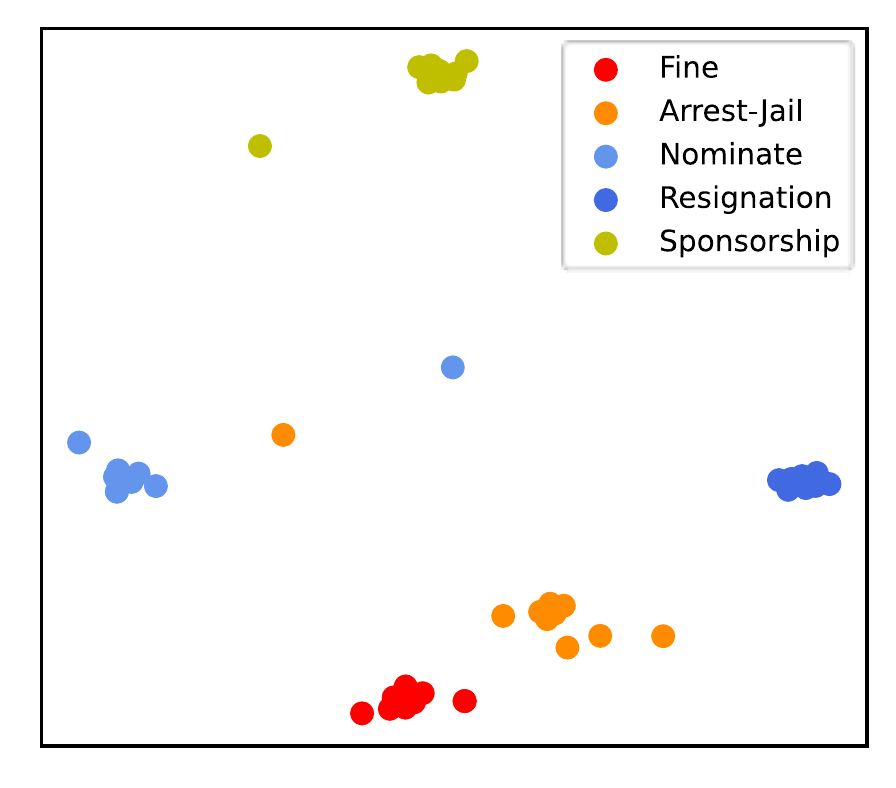}}
	\subfloat[Proto+PQCL]{\label{fig:visual_pqcl}\includegraphics[width=0.25\linewidth]{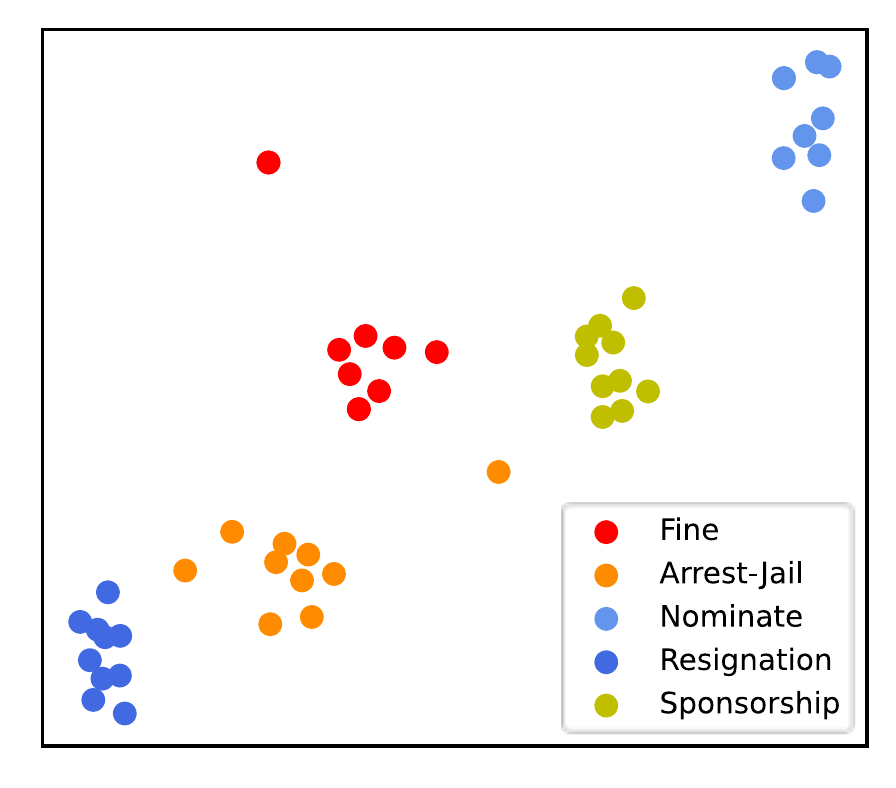}}
	\subfloat[Proto+HCL]{\label{fig:visual_hcl}\includegraphics[width=0.25\linewidth]{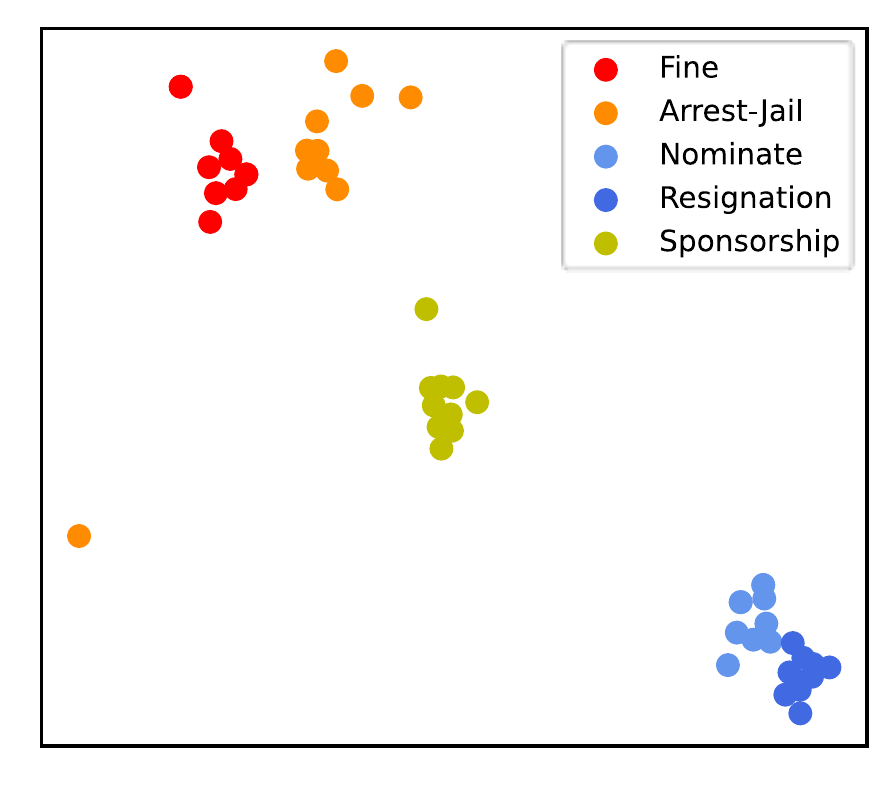}}\\
	\caption{Visualization of trigger embeddings in the same episode on FewEvent test set, under 5-way-10-shot setting. From left to right, the visualazation results of four FSED models are given respectively.}
	\label{fig:visual_all}
\end{figure*}

Given the same test episode, we use t-SNE to visualize the embeddings of triggers for the backbone model Proto with and without using contrast learning. From Figure~\ref{fig:visual_all}, we can observe that: (i) Proto achieves similar results as PA-CRF (see Figure~\ref{fig:visual_a}), with some instances hard to categorize. This proves that PA-CRF fails to improve representation learning by modeling label dependencies. (ii) SSCL and PQCL both contributes to the improvement of representations, while SSCL forms more compact clusters and PQCL learns more separable embedding space. (iii) When jointly using HCL, advantages of SSCL and PQCL are combined and more discriminative representations are produced to improve the performance. (iv) Besides, in Figure~\ref{fig:visual_hcl}, ``Fine'' and ``Arrest-Jail'', ``Nominate'' and ``Resignation'' have the common supertype ``Justice'' and ``Personnel'', respectively, and representations of event types belonging to the same supertype are relatively close. This shows that HCL automatically learns the relations between event types.



\section{Conclusion}


In this paper, we propose a contrastive learning method to make full use of limited data and produce more discriminative representations for FSED. Specifically, we first propose a hybrid contrastive loss, composed of support-support contrastive loss and prototype-query loss, to conduct supervised contrastive learning within support set and between support and query set respectively. Furthermore, we design a task-adaptive threshold method, to regularize the distance-based classifier in each episode. Experimental results show that by improving the representation learning and classifier learning simultaneously, our method boost the performance under all four settings on FewEvent dataset.
\section*{Limitations}

In FewEvent, each sentence has exactly one trigger word, so the sampling process is simplified into randomly selecting sentences containing specific events, and in each episode, it's unlikely for instances of ``O'' type to belong to any other event type that is not inluded in the current episode. However, in real-world scenarios, a sentence may contains multiple trigger words, which could bring more complicated settings. For example, we have to consider that instances of ``O'' type might belong to other event types that are not sampled in this episode, and the contrastive loss should be modified to adapt for such scenarios.

Besides, due to the huge memory cost, we only make full use of provided data, and do not consider data augmentation in contrastive learning, which has been proved effective in previous contrastive learning works. We believe that by conducting data augmentation and introducing more self-supervised signals, the performance of FSED could be further improved, which is worth for future research. 
\section*{Acknowledgements}

This work was supported in part by the National Natural Science Foundation of China under Grant No.62276110, Grant No.61602197, Grant No.61772076, in part by CCF-AFSG Research Fund under Grant No.RF20210005, and in part by the fund of Joint Laboratory of HUST and Pingan Property \& Casualty Research (HPL). The authors would also like to thank the anonymous reviewers for their comments on improving the quality of this paper.

\bibliography{main.bbl}
\bibliographystyle{acl_natbib}

\appendix

\section{Dataset Statistics}
\label{sec:A}

\begin{table}[htb]
\centering
\scalebox{1.0}{
\begin{tabular}{c|rrr}
    \toprule[1pt]
    \multicolumn{1}{c|}{Subset} & \#Class       & \#Trigger       & \#Avg.Len  \\
    \midrule[0.7pt]
    Train                 & 80    & 69088    & 36.5  \\
    Valid                      & 10       & 2274       & 38.6   \\
    Test                      & 10    & 748    & 30.8 \\
    \bottomrule[1pt]
\end{tabular}
}
\caption{The statistics of FewEvent Dataset. \#Class, \#Trigger and \#Avg.Len denotes the number of classes, the number of triggers and the average length of sentences in each split part respectively.}
\label{table_statistics}
\end{table}

\begin{figure}[htb]
    \centering
    \includegraphics[width=0.48\textwidth]{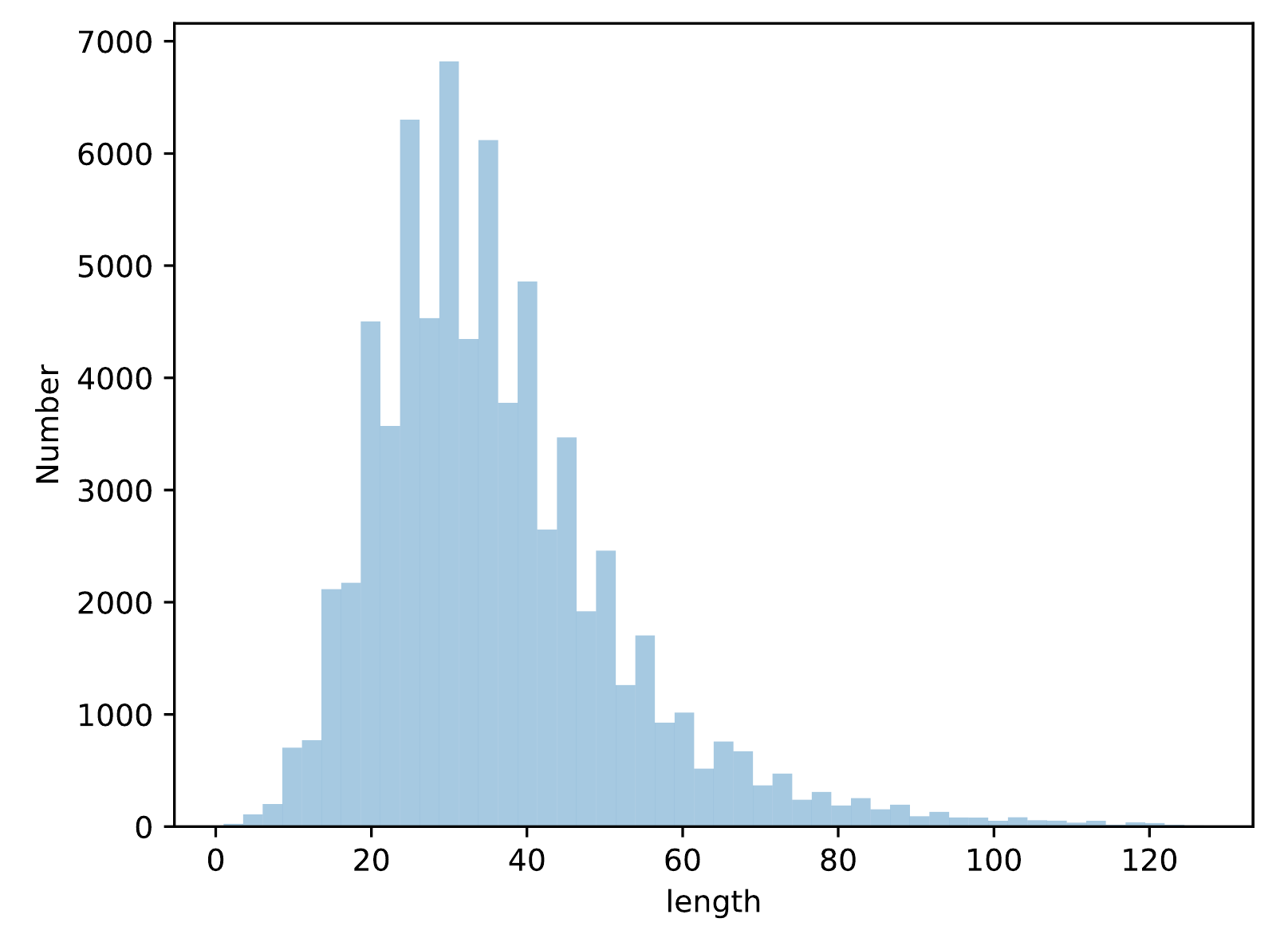}
    \caption{Length distribution of sentences in FewEvent dataset.}
    \label{fig:length_distribution}
\end{figure}

The statistics of FewEvent dataset are listed in Table~\ref{table_statistics}, including the number of event types (\#Class), the number of triggers (\#Trigger) and the average length of sentences (\#Avg.Len) of train, valid and test set. The length distribution of sentences in FewEvent is illustrated in Figure~\ref{fig:length_distribution}, showing that most of the sentence lengths are within 128. Thus we set the maximum length to 128.

\section{Parameter Study}
\label{sec:C}
\begin{figure}[htb]
    \centering
    \includegraphics[width=1.0\linewidth]{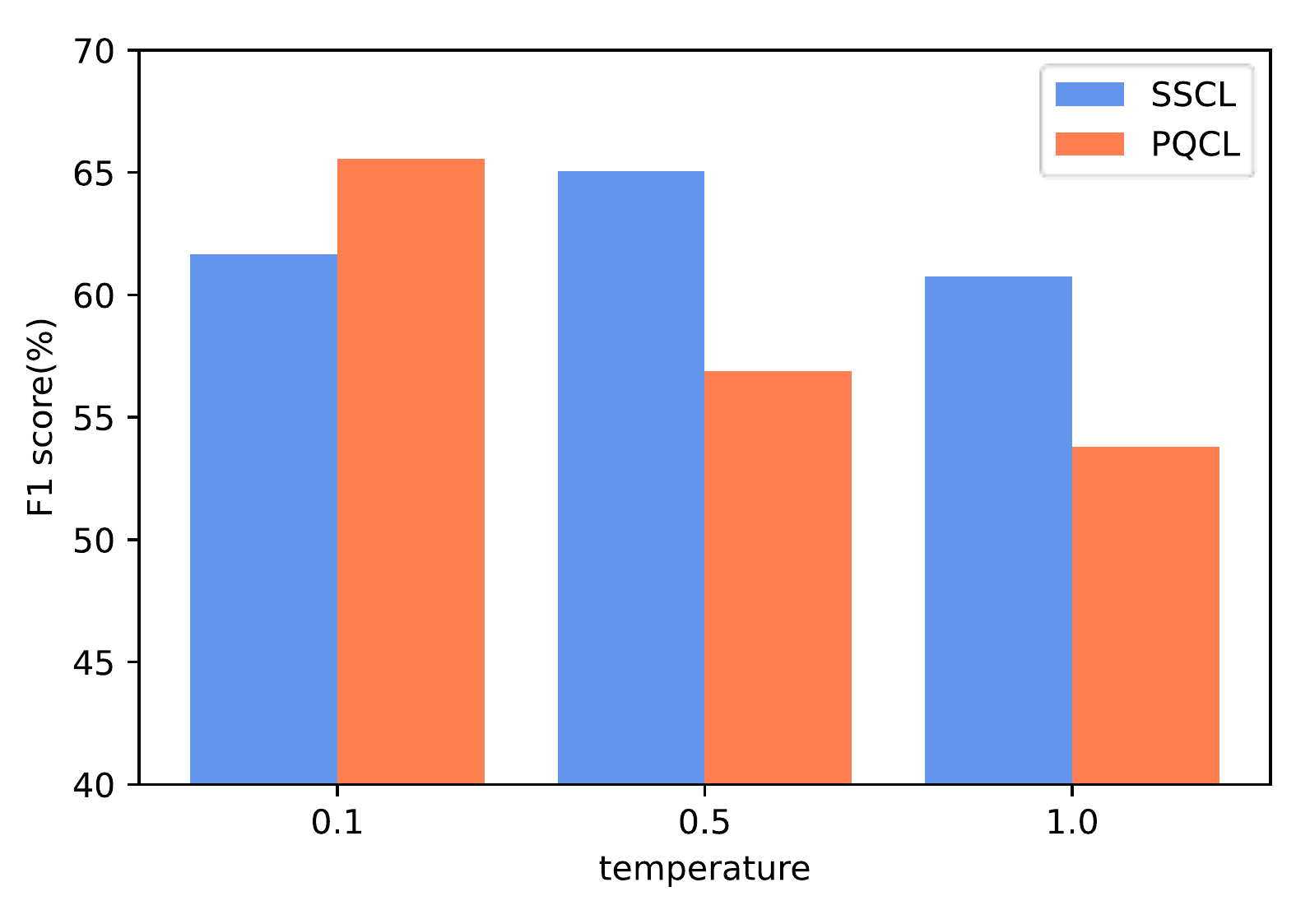}
    \caption{F1 scores($10^{-2}$) over different temperature values on the two contrastive losses. The results are obtained under 5-way-5-shot setting in FewEvent test set.}
    \label{fig:parameter_study}
\end{figure}

We study the influence of the scalar temperature $\tau$ on our two contrastive losses, SSCL and PQCL. Figure~\ref{fig:parameter_study} shows the results of the backbone model with only SSCL or PQCL, respectively. We can empirically observe that, PQCL benefits from smaller temperature ($\tau = 0.1$) and a larger temperature brings improvement for SSCL ($\tau = 0.5$).

\section{Bottleneck Analysis for Query-anchored Cross-entropy Loss}
\label{sec:D}
\begin{theorem}
\label{proposition-obj}
  Optimized with Eq. (\ref{eq:ce_loss}) could not produce discriminative representations for prototypes, and further harms the representation learning of anchor query instance.
\end{theorem}
\vspace{-1ex}
\begin{proof}
For simplicity, we assume dot product is used as the similarity metric in Eq. (\ref{eq:ce_loss}). The loss $\mathcal{L}_{CE}$ can thus be converted as follows:
\begin{equation}
    \begin{aligned}
        \mathcal{L}_{CE} & = -\operatorname{log}\frac{\operatorname{exp}(\vec{h}_i \cdot \vec{p}^{pos})}{\operatorname{exp}(\vec{h}_i \cdot \vec{p}^{pos}) + \sum_{n}\operatorname{exp}(\vec{h}_i \cdot \vec{p}^{n})}\\
        & = -\operatorname{log}\frac{1}{1 + \sum_{n}\frac{\operatorname{exp}(\vec{h}_i \cdot \vec{p}^n)}{\operatorname{exp}(\vec{h}_i \cdot \vec{p}^{pos})}}\\
        & = \operatorname{log}(1 + \sum_{n}\frac{\operatorname{exp}(\vec{h}_i \cdot \vec{p}^n)}{\operatorname{exp}(\vec{h}_i \cdot \vec{p}^{pos})})\\
        & = \operatorname{log}(1 + \sum_{n}\operatorname{exp(\vec{h}_i \cdot \vec{p}^n - \vec{h}_i \cdot \vec{p}^{pos})}),
    \end{aligned}
\end{equation}
where $\vec{h}_i$ is the representation of the anchor query instance, $\vec{p}^{pos}$ is the positive prototype of the same class, and $\vec{p}^n$ is a negative prototype of different classes. Then we calculate the partial derivative with respect to $\vec{h}_i$ to compute the gradient of $\vec{h}_i$. Let $\Delta_n = \operatorname{exp(\vec{h}_i \cdot \vec{p}^n - \vec{h}_i \cdot \vec{p}^{pos})}$, the calculation procedure is as follows:
\begin{equation}
    \label{eq:analysis}
    \begin{aligned}
        \frac{\partial \mathcal{L}_{CE}}{\partial \vec{h}_i} & = \frac{\sum_{n}\operatorname{exp(\vec{h}_i \cdot \vec{p}^n - \vec{h}_i \cdot \vec{p}^{pos})} |_{\vec{h}_i}}{1 + \sum_{n}\operatorname{exp(\vec{h}_i \cdot \vec{p}^n - \vec{h}_i \cdot \vec{p}^{pos})}}\\
        & = \frac{\sum_{n}\operatorname{exp(\vec{h}_i \cdot \vec{p}^n - \vec{h}_i \cdot \vec{p}^{pos})} (\vec{p}^n - \vec{p}^{pos})}{1 + \sum_{n}\operatorname{exp(\vec{h}_i \cdot \vec{p}^n - \vec{h}_i \cdot \vec{p}^{pos})}}\\
        & = \frac{\sum_{n}\Delta_n (\vec{p}^n - \vec{p}^{pos})}{1 + \sum_{n}\Delta_n}.
    \end{aligned}
\end{equation}

Similarly, we can compute the gradient of $\vec{p}^{n}$ and $\vec{p}^{pos}$ as follows:
\begin{gather}
    \frac{\partial \mathcal{L}_{CE}}{\partial \vec{p}^{n}} = \frac{\Delta_n \vec{h}_{i}}{1 + \sum_{n}\Delta_n},\\
    \frac{\partial \mathcal{L}_{CE}}{\partial \vec{p}^{pos}} = -\frac{\sum_{n}\Delta_n \vec{h}_{i}}{1 + \sum_{n}\Delta_n}.
\end{gather}

We can conclude that the update direction of $\vec{h}_i$ is better than $\vec{p}^{pos}$ and $\vec{p}^{n}$, so it's hard for prototypes to learn discriminative representations. Furthermore, if $\vec{p}^{pos}$ and $\vec{p}^{n}$ are close, then the gradient of $\vec{h}_i$ approaches to zero, making it hard to learn discriminative representation for the anchor query instance as well.
\end{proof}

Theorem~\ref{proposition-obj} indicates that bad prototype representations bring bad query representations as well, and eventually harm the performance of FSED models. Therefore, we propose the contrastive learning based method to learn more discriminative representations for both support set and query set.


\end{document}